\documentclass[lettersize,journal]{IEEEtran}
\usepackage[utf8]{inputenc}
\usepackage{amsmath,amsfonts,amssymb}
\usepackage{algorithmic}
\usepackage{algorithm}
\usepackage{array}
\usepackage[caption=false,font=normalsize,labelfont=sf,textfont=sf]{subfig}
\usepackage{textcomp}
\usepackage{stfloats}
\usepackage{url}
\usepackage{verbatim}
\usepackage{graphicx}
\usepackage[numbers]{natbib}
\usepackage{booktabs}
\usepackage{multirow}
\usepackage{makecell}
\usepackage{bbding}
\usepackage{xcolor}
\usepackage[switch,columnwise]{lineno}
\usepackage[colorlinks]{hyperref}
\usepackage[dvipsnames]{xcolor}
\usepackage[table]{xcolor}
\usepackage{geometry}
\usepackage{amssymb}
\usepackage[table]{xcolor} 
\definecolor{lightyellow}{HTML}{FFFDE7} 
\definecolor{lightblue}{HTML}{E3F2FD}  
\usepackage[table,xcdraw]{xcolor}
\usepackage[table]{xcolor} 
\usepackage{colortbl}
\usepackage{xcolor}
\usepackage{tabularx} 

\begin{document}

\title{Hyperbolic Geometry for Open-World Object Detection in Remote Sensing Imagery}

\author{
Wuzhou~Li,
Jiawei~Zhou,
Shenghang~Wang,
and~Xiang~Li

\thanks{ \textit{(Corresponding author: .)}}
\thanks{Wuzhou~Li is with the School of Computer Science and Artificial Intelligence, Wuhan Textile University, Wuhan 430072, China  (email: 2025054@wtu.edu.cn).}%
\thanks{Jiawei~Zhou is with the Electronic Information School, Wuhan University, Wuhan 430072, China (email: zhoujw@whu.edu.cn).}%
\thanks{Shenghang~Wang is with the Electrical and Computer Engineering, Ohio State University, Columbus, OH, USA (email: wang.19780@osu.edu).}%
\thanks{Xiang~Li is with the School of Artificial Intelligence, Wuhan University, Wuhan 430072, China (email: xiangli92@ieee.org).}
}

\markboth{Journal of \LaTeX\ Class Files,~Vol.~14, No.~8, August~2021}%
{Li \MakeLowercase{\textit{et al.}}: Hyperbolic Geometry for Open-World Object Detection in Remote Sensing Imagery}


\maketitle

\begin{abstract}
Open-world object detection (OWOD) extends closed-set detection by requiring models to identify unknown objects and incrementally learn them once annotations become available. In remote sensing imagery, object categories often exhibit latent hierarchical relationships that may be inadequately represented in the Euclidean spaces commonly adopted by existing methods, limiting unknown-object recall and incremental-learning performance. To address this issue, we investigate hyperbolic geometry for OWOD in remote sensing imagery and propose HyRS-OWOD. To improve unknown object recall, we design a two-step unknown-object discovery mechanism: a Decoupled Objectness Learning (DOL) module that disentangles foreground perception from semantic information to separate foreground proposals from background regions, followed by a Hyperbolic Uncertainty Learning (HUL) component that leverages the radius of hyperbolic embeddings as an uncertainty-aware cue for known–unknown discrimination. For incremental learning, we develop a Hyperbolic Metric Learning (HML) strategy that enhances inter-class separability, facilitating the incorporation of novel categories while mitigating catastrophic forgetting. Experiments on three remote sensing benchmarks demonstrate consistent improvements in unknown recall and incremental learning over state-of-the-art OWOD methods. 
\end{abstract}


\section{Introduction}

\begin{figure}[t]
\centering
\includegraphics[width=\columnwidth]{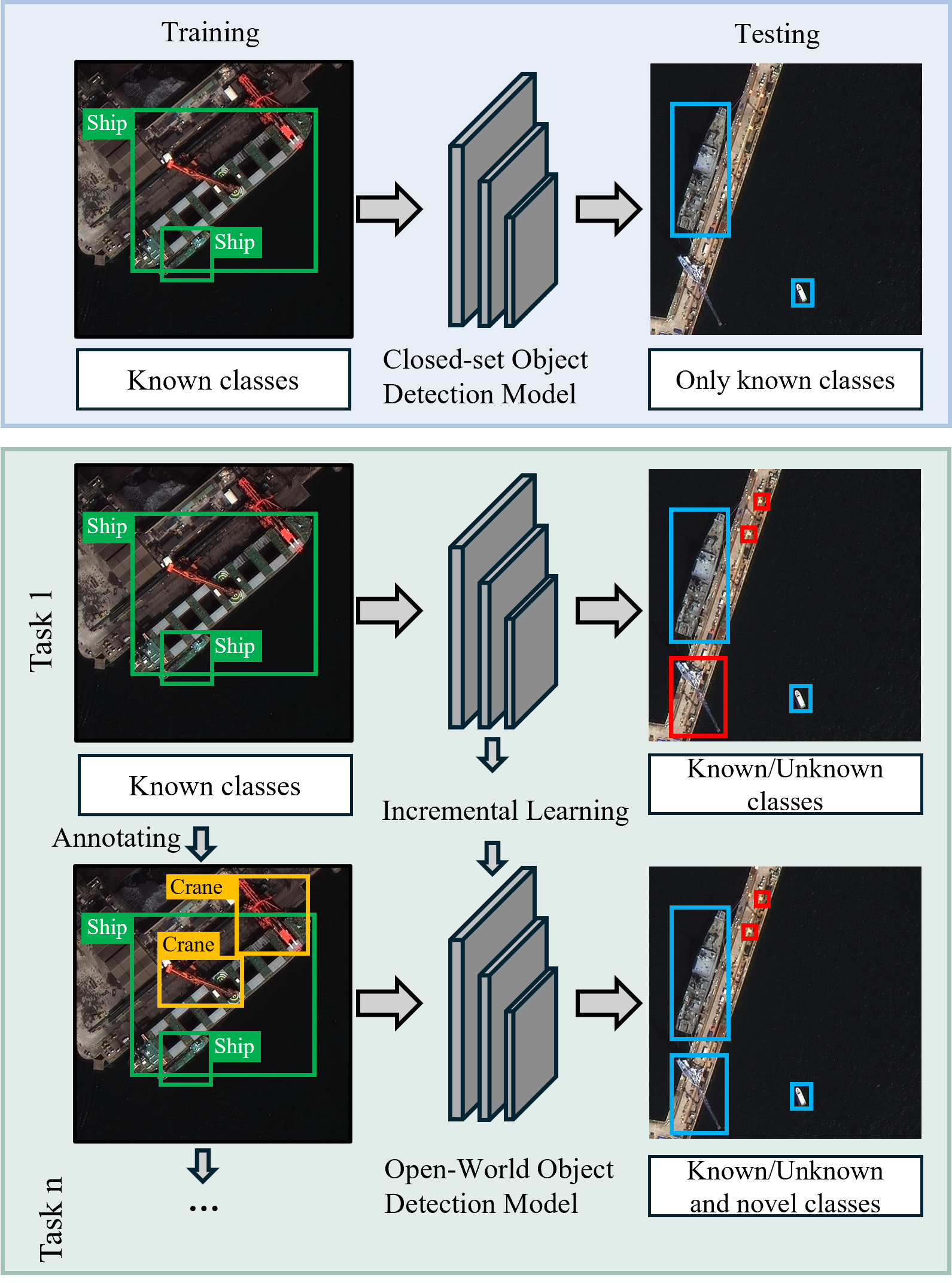}
\caption{Comparison between closed-set and open-world object detection in remote sensing imagery. Closed-set detectors recognize objects from only a predefined set of known categories, whereas OWOD additionally identifies unseen objects as unknown and incrementally learns them as novel classes once annotations become available.}
\label{fig1}
\end{figure}

\IEEEPARstart{O}{bject} detection is a fundamental computer vision task that underpins a wide range of remote sensing applications, including urban planning, environmental monitoring, and disaster management \cite{1-li2020object,2-zhang2023remote}. To operate reliably in dynamic real-world scenarios, object detectors must be able to handle unknown objects that continuously emerge and may contain valuable information. However, conventional closed-set detectors are typically trained on a predefined set of categories and are expected to recognize only objects belonging to these known (labeled) classes, whereas unknown (unlabeled) objects are often misclassified as similar known categories or ignored as background, limiting the applicability of such detectors in open-world environments. In response, open-world object detection (OWOD) has emerged as a promising paradigm to bridge this gap~\cite{3-joseph2021towards,4-li2024open}. As illustrated in Fig.~\ref{fig1}, OWOD requires a detector not only to recognize known objects but also to identify previously unseen objects as unknown and incrementally incorporate newly annotated categories across successive learning stages. Recently, OWOD has attracted growing attention in the remote sensing field~\cite{6-lang2024toward,7-fang2024open}, as exemplified by the study of Tan et al.~\cite{8-Tan2024OPODet}, which extend the unknown-aware region proposal network to handle oriented unknown objects in remote sensing imagery. Hu et al.~\cite{5-Hu2026Unveiling} further leverage the Segment Anything Model (SAM) to discover potential unknown objects and employ a geometric-mean-based thresholding function to reduce the influence of noisy labels in the raw SAM outputs. 

Despite these advances, OWOD in remote sensing imagery remains challenging due to two major issues: the low recall of unknown objects and catastrophic forgetting of previously learned classes during incremental learning. The first issue stems primarily from the lack of supervision, as unknown objects, unlike known classes, are unlabeled during training and can therefore be easily confused with background regions. Furthermore, in remote sensing imagery, unknown objects may correspond to fine-grained extensions or semantic subcategories of known classes, making them prone to confusion with similar known objects, which further suppresses unknown-object recall. The second issue is mainly rooted in high inter-class similarity: the continuous introduction of novel classes shifts the input data distribution, leading to catastrophic forgetting of previously learned classes during adaptation~\cite{9-gepperth2016incremental}. This effect is especially pronounced when novel classes are visually similar to base classes or correspond to fine-grained variants, as their feature distributions may significantly overlap with those of previously learned classes, thereby exacerbating the problem.

To tackle these challenges, our key insight is to embed the underlying hierarchical structure of remote sensing categories into hyperbolic space. Although Euclidean space has long been the de facto manifold for visual representation learning, its flat geometry cannot faithfully preserve hierarchical or tree-like relationships without considerable distortion. Remote sensing categories, however, often exhibit such hierarchical structure, where unknown objects may be semantically related to known categories at different levels of granularity. For example, a new type of destroyer can be regarded as a type of warship, while warship itself is a subcategory of ship. Embedding such data in Euclidean space may distort the underlying semantic relationships, leading to suboptimal performance. In contrast, hyperbolic space, a Riemannian manifold endowed with constant negative curvature, offers a more suitable geometry for modeling the hierarchical structure of remote sensing data. More importantly, its geometric properties are relevant to the two principal challenges of OWOD. On one hand, the hyperbolic radius of embeddings can naturally serve as a measure of uncertainty for discovering potential unknown objects~\cite{10-Khrulkov2020Hyperbolic,11-hong2023curved}. On the other hand, its exponential volume growth and hierarchical structure provide a suitable foundation for metric learning, helping maintain discriminative relationships among previously learned and novel classes during incremental learning, thereby mitigating catastrophic forgetting.

Motivated by these observations, we introduce hyperbolic geometry into OWOD for remote sensing imagery. To improve the recall of unknown objects, we design a two-step unknown-object discovery mechanism comprising foreground–background separation and known–unknown discrimination. In the first step, we propose Decoupled Objectness Learning (DOL), which explicitly models class-agnostic objectness to distinguish foreground proposals from background regions. Although objectness is intended to be class-agnostic, supervision solely from known classes may introduce a confounding effect that suppresses potential unknown objects~\cite{31-Sun2024Orthogonality,32-wang2023random,58-Pourhoseingholi2012HowTC}. To alleviate this issue, DOL introduces a decoupling loss that reduces the statistical correlation between objectness scores and class predictions. In the second step, termed Hyperbolic Uncertainty Learning (HUL), we employ an uncertainty-based heuristic to discriminate unknown objects from known classes. Intuitively, among foreground objects, unknowns can be regarded as outliers with high predictive uncertainty. Inspired by prior studies that leverage uncertainty estimation for out-of-distribution classification~\cite{11-hong2023curved} and segmentation tasks~\cite{12-chen2023hyperbolic}, HUL uses the hyperbolic radius of a learned embedding, measured by its distance to the origin, as a geometry-aware uncertainty cue: proposals with smaller radii tend to correspond to more abstract and ambiguous concepts. As such, foreground proposals with high objectness scores but small hyperbolic radii are more likely to be identified as unknown objects.

To address catastrophic forgetting, our key idea is to learn stable and discriminative representations. Metric learning has been an effective method for learning an embedding space or distance metric in which samples from the same class are drawn closer together, while samples from different classes are pushed farther apart~\cite{13-Tran2020HyperML,14-cui2023rethinking}. Since remote sensing categories typically exhibit a hierarchical structure, we further perform metric learning in hyperbolic space. Compared with its Euclidean counterpart, hyperbolic space exhibits exponential volume growth, which naturally accommodates hierarchical trees with low distortion~\cite{10-Khrulkov2020Hyperbolic}. When combined with metric learning, this geometry promotes more compact intra-class representations and larger inter-class margins. Accordingly, we propose Hyperbolic Metric Learning (HML), which combines a class-balanced proposal buffer with distance-based hard-negative weighting. The proposal buffer retains representative embeddings of previously learned classes, while the weighting mechanism emphasizes visually confusing proposals from different classes. Consequently, HML reduces interference between previously learned and novel classes, thereby mitigating catastrophic forgetting during incremental learning.

To summarize, our main contributions are as follows:
\begin{enumerate}

    \item We propose HyRS-OWOD, a novel open-world object detection (OWOD) framework for remote sensing imagery that, to the best of our knowledge, is the first to investigate the effectiveness of hyperbolic geometry for OWOD in the remote sensing domain. Our method enables the detection of known classes, discovery of unknown objects, and incremental learning of novel classes in a unified manner.
  
    \item We design a two-step unknown-object discovery mechanism comprising Decoupled Objectness Learning (DOL), which models class-agnostic objectness with a decoupling loss to separate foreground from background, and Hyperbolic Uncertainty Learning (HUL), which leverages hyperbolic uncertainty to identify unknown objects from known categories, thereby improving unknown-object recall.

    \item We propose Hyperbolic Metric Learning (HML) for incremental object detection. By combining a class-balanced proposal buffer with distance-based hard-negative weighting, HML promotes intra-class compactness and inter-class separation while reducing interference between previously learned and novel classes.
    
    \item Comprehensive experiments on three remote sensing object detection benchmarks demonstrate the effectiveness and superiority of our proposed HyRS-OWOD over existing approaches.
\end{enumerate}

\section{Related Work} 
\subsection{Object Detection}

Deep learning has substantially advanced object detection through representative architectures such as the YOLO family~\cite{15-Redmon2016YOLO,16-Redmon2017YOLO9000,17-Redmon2018Yolov3,18-Bochkovskiy2020Yolov4}, two-stage detectors represented by Faster R-CNN~\cite{19-Girshick2014RCNN,20-Ren2015FasterRCNN,21-Lin2017FPN}, and transformer-based detectors such as DETR~\cite{22-carion2020end,23-zhu2020deformable}. In remote sensing, these architectures have been adapted to address domain-specific challenges, including small objects~\cite{24-hua2025survey}, complex backgrounds~\cite{2-zhang2023remote}, and arbitrarily oriented objects~\cite{25-wen2023comprehensive}. Nevertheless, most existing methods are developed under a static, closed-set assumption, restricting their detection to a fixed set of predefined classes. To relax this assumption, open-set detection aims to identify objects outside the known label space~\cite{6-lang2024toward}, whereas open-vocabulary detection exploits external semantic knowledge to recognize a broader range of categories~\cite{26-pan2025locate}. These developments highlight the unique challenges of open-world perception and motivate the need for a more continual and comprehensive approach. 

\subsection{Open-World Object Detection}
OWOD combines the detection of unknown objects with their incremental learning as novel classes once annotations are progressively provided. Joseph et al.~\cite{3-joseph2021towards} first formulated this task and proposed ORE,  a Faster R-CNN-based framework that employs an unknown-aware pseudo-labeling scheme and an energy-based classifier to distinguish unknown objects from known classes. Following this seminal work, several studies have adopted different heuristic assumptions for discovering potential unknown objects. OW-DETR~\cite{27-Gupta2022OW-DETR}, for example, employs attention-driven pseudo-labeling to mine potential unknown samples. Wang et al.~\cite{32-wang2023random} utilize random proposal generation to mitigate the confounding effect and explore more potential proposals of unknown objects. CAT~\cite{28-Ma2023CAT} decouples localization and classification with a cascade decoder and proposes a self-adaptive pseudo-labeling scheme for generating robust pseudo-labels. Sun et al.~\cite{31-Sun2024Orthogonality} further address the interference between objectness and class information via orthogonalization. Zohar et al.~\cite{30-Zohar2023PROB} directly learn a probabilistic objectness head from only known classes, eliminating the need for pseudo-labels. Another line of work introduces foundation-model and vision-language models~\cite{33-Zohar2023OpenWO,34-He2025Recalling,35-Xi2024UMB,36-xi2025owvap,37-li2026openvocabularyopenworld} into OWOD for external semantic knowledge, leveraging vision-language alignment, textual descriptions, and attribute-level cues to enhance generalization to unseen categories, partially bridging OWOD with open-vocabulary object detection. 

Compared with generic-domain OWOD, research on OWOD in remote sensing remains relatively limited. OPODet~\cite{8-Tan2024OPODet} primarily improves detection of unknown rotated objects in remote sensing images. Saini et al.~\cite{38-saini2025advancing} employ multimodal large language models to assign semantic labels to unknown candidate regions generated by closed-set detectors, enabling the automatic discovery and naming of unknown classes. However, these methods focus mainly on unknown-object discovery without subsequent incremental learning, thus falling short of the full OWOD requirements. More recently, Hu et al.~\cite{5-Hu2026Unveiling}  proposed a SAM-guided OWOD framework with a label-mapping alignment mechanism to filter noisy candidate proposals. Despite this progress, existing remote sensing OWOD methods still largely overlook the latent hierarchical structure among remote sensing categories. Motivated by these observations, we explore hyperbolic geometry for OWOD in remote sensing images, aiming to better model the semantic relationships between unknown objects and known classes.

\subsection{Hyperbolic Geometry}
Hyperbolic geometry has attracted increasing attention in representation learning because of its ability to represent hierarchical and tree-like structures with low distortion~\cite{54-Mettes2024Hyperbolic,55-Peng2022Hyperbolic}. Hyperbolic representations have been applied to natural language processing~\cite{39-Ganea2018Hyperbolic}, knowledge graph modeling~\cite{40-WANG2023HyGGE}, and graph neural networks~\cite{41-Chami2019Hyperbolic}. More recently, they have been extended to various computer vision tasks, including continual learning~\cite{42-Gao2023Exploring,14-cui2023rethinking}, few-shot learning~\cite{43-Zhang2022Hyperbolic}, image classification~\cite{44-Guo2022Clipped}, semantic segmentation~\cite{45-Chen2024Hyperbolic,46-Atigh2022Hyperbolic}, and object detection~\cite{47-Lang2022On}. These studies suggest that hyperbolic embeddings can effectively capture latent semantic similarities and hierarchical relationships in visual data, providing a complementary alternative to conventional Euclidean representations. Moreover, another key property lies in the Poincaré model of hyperbolic spaces, where the embedding radius—i.e., the distance to the origin—can serve as a measure of model uncertainty~\cite{39-Ganea2018Hyperbolic,54-Mettes2024Hyperbolic}. This property has been exploited in tasks such as visual anomaly recognition~\cite{11-hong2023curved} and image segmentation~\cite{45-Chen2024Hyperbolic}.

Hyperbolic geometry has also recently been introduced into OWOD. Kong et al.~\cite{48-kong2024hyperbolic} develop a hyperbolic learning framework that models hierarchical relationships between visual and caption embeddings, thereby reducing the influence of hallucinated synthetic captions. Doan et al.~\cite{49-doan2024hyp} propose Hyp-OW, which combines hyperbolic contrastive learning for object representation, a superclass regularizer for modeling class-level hierarchies, and adaptive relabeling for retrieving unknown objects based on hyperbolic distance. These studies demonstrate the potential of hyperbolic geometry for modeling category hierarchies and facilitating unknown-object discovery. However, they are primarily designed for natural-scene imagery, and the application of hyperbolic geometry to remote sensing OWOD remains underexplored. 

\subsection{Hyperbolic Learning in Remote Sensing}
Remote sensing imagery often contains semantically related object categories with implicit hierarchical structures, which naturally motivates the use of hyperbolic spaces. Recent studies have introduced hyperbolic geometry into remote sensing tasks, including image classification~\cite{50-HAMZAOUI2024Hyperbolic}, semantic segmentation~\cite{51-Li2022Hybridizing,52-CUI2024Representation}, and change detection~\cite{53-Yang2024Hyperboloid}. In this work, we introduce hyperbolic geometry into remote sensing OWOD to better capture the latent hierarchical relationships between known and unknown classes and formulate an uncertainty-aware unknown object discovery strategy based on the hyperbolic radius. Moreover, we further perform metric learning in hyperbolic space to preserve intra-class compactness and inter-class separability during incremental learning, thereby advancing OWOD for remote sensing imagery. 

\section{Preliminaries}
In this section, we begin by introducing the fundamentals of hyperbolic space, followed by a formal definition of the Poincaré ball model. We then describe several commonly used operations within this model. 

\subsection{Hyperbolic Space}
Hyperbolic space is defined as a complete, simply connected Riemannian manifold with constant negative sectional curvature. Among its various isometric models, the Poincaré ball model has conformal properties, which enable effective modeling of hierarchical data structures and facilitate efficient optimization. 
\subsection{Poincaré Ball Model}
An $n$-dimensional Poincaré ball model with constant negative curvature $-c$ $(c>0)$ is defined as a Riemannian manifold $\mathbb{M}_{c}^{n} = (\mathcal{B}_{c}^{n}, g^{\mathbb{B}})$, where $\mathcal{B}_{c}^{n}$ denotes an open ball embedded in the Euclidean space $\mathbb{R}^{n}$:
\begin{equation}
\mathcal{B}_{c}^{n} = \left\{ \mathbf{x} \in \mathbb{R}^{n}: c\|\mathbf{x}\|^{2} < 1 \right\},
\end{equation}
with $\|\cdot\|$ the Euclidean norm. The radius of the ball is $1/\sqrt{c}$. The Riemannian metric tensor of the Poincaré ball model is conformal to the Euclidean metric and is given by
\begin{equation}
g^{\mathbb{B}}_{\mathbf{x}} = \left(\lambda_{\mathbf{x}}^{c}\right)^{2} g^{E},
\end{equation}
where $g^{E}$ denotes the Euclidean metric tensor and $\lambda_{\mathbf{x}}^{c}$ is the conformal factor:
\begin{equation}
\lambda_{\mathbf{x}}^{c} = \frac{2}{1 - c\|\mathbf{x}\|^{2}}.
\end{equation}
As $\|\mathbf{x}\|$ approaches the boundary, the conformal factor grows without bound, enabling the model to capture hierarchical structures with exponentially expanding capacity near the boundary.
\subsection{M\"obius addition}
We adopt the formalism of M\"obius gyrovector spaces, which generalize several Euclidean operations to hyperbolic geometry and enables algebraic computations within the Poincaré ball model. 
For $\mathbf{x}, \mathbf{y} \in \mathcal{B}_c^n$, the M\"obius addition $\oplus_c$ is defined as:
\begin{equation}
\mathbf{x} \oplus_c \mathbf{y} =
\frac{
\left(1 + 2c\langle \mathbf{x}, \mathbf{y} \rangle + c\|\mathbf{y}\|^2\right)\mathbf{x} +
\left(1 - c\|\mathbf{x}\|^2\right)\mathbf{y}
}{
1 + 2c\langle \mathbf{x}, \mathbf{y} \rangle + c^2\|\mathbf{x}\|^2 \|\mathbf{y}\|^2
}.
\end{equation}

\subsection{Exponential Map}
For $\mathbf{x} \in \mathcal{B}_c^n$ and $\mathbf{v} \in T_{\mathbf{x}}\mathcal{B}_c^n$, the exponential map is defined as:
\begin{equation}
\exp_{\mathbf{x}}^c(\mathbf{v}) = 
\mathbf{x} \oplus_c \left( 
\tanh\left( \sqrt{c} \frac{\lambda_{\mathbf{x}}^c \|\mathbf{v}\|}{2} \right) 
\frac{\mathbf{v}}{\sqrt{c} \|\mathbf{v}\|} 
\right).
\end{equation}

\subsection{Distance and Hyperbolic Radius}
For $\mathbf{x}, \mathbf{y} \in \mathcal{B}_c^n$, the hyperbolic distance is given by
\begin{equation}
d_c(\mathbf{x}, \mathbf{y}) = \frac{1}{\sqrt{c}} \operatorname{arccosh}\left( 
1 + \frac{2c\|\mathbf{x} - \mathbf{y}\|^2}{(1 - c\|\mathbf{x}\|^2)(1 - c\|\mathbf{y}\|^2)} 
\right).
\end{equation}

The hyperbolic radius of a point $\mathbf{x}$ is defined as its Poincaré distance to the ball origin:
\begin{equation}
r_c(\mathbf{x}) \triangleq d_c(\mathbf{x}, \mathbf{0}) = \frac{2}{\sqrt{c}} \operatorname{arctanh}\left( \sqrt{c}\|\mathbf{x}\| \right).
\end{equation}

The hyperbolic radius of an object proposal serves as a measure of model uncertainty: higher confidence corresponds to embeddings lying farther away from the origin in hyperbolic space, whereas lower confidence leads to embeddings located closer to the origin~\cite{10-Khrulkov2020Hyperbolic,12-chen2023hyperbolic}. 

\subsection{\texorpdfstring{$\delta$}{delta}-Hyperbolicity}
Following the analysis in~\cite{10-Khrulkov2020Hyperbolic}, we compute the Gromov \texorpdfstring{$\delta$}{delta}-hyperbolicity of each remote sensing dataset to measure its structural properties. This evaluation is performed by first computing the Gromov product for points $x, y, z \in \mathcal{X}$, which is defined as:

\begin{equation}
(y, z)_x = \frac{1}{2} \left( d(x, y) + d(x, z) - d(y, z) \right),
\end{equation}
where $(\mathcal{X}, d)$ denotes an arbitrary metric space. For a set of points, we construct the matrix $M$ of pairwise Gromov products. The $\delta$-hyperbolicity is then computed as:

\begin{equation}
\delta = \max_{i,j} \left[ (M \otimes M)_{ij} - M_{ij} \right],
\end{equation}
where $\otimes$ denotes the min-max matrix product defined as:

\begin{equation}
(A \otimes B)_{ij} = \max_{k} \min \{ A_{ik}, B_{kj} \}.
\end{equation}
with $i, j, k$ denoting the indices of matrices $A$ and $B$.

We further compute the relative hyperbolicity to ensure scale invariance. Its value ranges from 0 to 1, where values closer to 0 indicate stronger hierarchical structure in the data, while values closer to 1 suggest a weaker hierarchy.

\begin{equation}
\delta_{\mathrm{rel}} = \frac{\delta}{\max_{i,j} M_{ij}}.
\end{equation}
As shown in~\cite{10-Khrulkov2020Hyperbolic}, $\delta_{\mathrm{rel}}$ can be used to estimate the manifold curvature of the Poincaré ball model:

\begin{equation}
c(\mathcal{X}) = \left(\frac{0.144}{\delta_{\mathrm{rel}}}\right)^2.
\end{equation}

We follow the above procedure to evaluate $\delta_{\mathrm{rel}}$ on remote sensing detection datasets, including NWPU VHR-10, DIOR, and DOTA. For each dataset, we extract region-level embeddings from object proposals using different detection models. We randomly sample 450 embeddings per run and the final results are reported in Table~\ref{tab1}.

\begin{table}[t]
\centering
\caption{Relative $\delta$-hyperbolicity values of embeddings extracted by different feature extractors. }
\label{tab1}
\begin{tabular}{lccc}
\toprule
 & NWPU VHR-10 & DIOR & DOTA  \\
\midrule
VGG16  & 0.2306 & 0.2296 & 0.2525  \\
GoogLeNet & 0.2196 & 0.2415 & 0.2372  \\
ResNet50   & 0.2544 & 0.2751 & 0.2910 \\
\bottomrule
\end{tabular}
\end{table}

\section{Methodology}
\subsection{Problem Formulation}
Let $\mathcal{K}^{t} = \{1, 2, \ldots, C\}$ denote the annotated known classes, and $\mathcal{U}^{t} = \{C+1,\ldots\}$ denote the unbounded set of unknown classes of interest that may appear during inference. Given a time step $t$, the known object classes $\mathcal{K}^{t}$ are labeled in the dataset $\mathcal{D}^{t} = \{\mathcal{I}^{t}, \mathcal{L}^{t}\}$, which consists of $N$ input images $\mathcal{I}^{t} = \{I_{1}, \ldots, I_{N}\}$ and their corresponding label sets $\mathcal{L}^{t} = \{L_{1}, \ldots, L_{N}\}$. Each $L_i=\{l_{i1},\ldots,l_{iP_i}\}$ contains the annotations of $P_i$ object instances in image $I_i$. Each instance annotation is represented as $l_{ij}=[c_{ij},x_{ij},y_{ij},w_{ij},h_{ij}]$, where $c_{ij}$ denotes its class label,
$(x_{ij},y_{ij})$ denotes the center coordinates of its bounding box, and $w_{ij}$ and $h_{ij}$ denote its width and height, respectively.

In OWOD, the trained model $\mathcal{M}^t$ is required to detect objects from known classes $\mathcal{K}^{t}$ and identify unknown objects belonging to $\mathcal{U}^{t}$, labeling them as "unknown". The model then sends the discovered unknown objects to an oracle, which annotates the new classes of interest. Together with their corresponding training examples, these annotations update $\mathcal{D}^{t}$ to $\mathcal{D}^{t+1}$ and $\mathcal{K}^{t}$ to $\mathcal{K}^{t+1} = \{1, 2, \ldots, C, \ldots, C+n\}$. The model incorporates these $n$ novel classes into the set of known classes and incrementally updates itself to $\mathcal{M}^{t+1}$ without retraining from scratch on the entire dataset $\mathcal{D}^{t+1}$ while preserving previously learned knowledge. This cycle continues over the model's lifespan.

\subsection{Overall Architecture}

\begin{figure*}[t]
    \centering
    \includegraphics[width=0.98\textwidth]{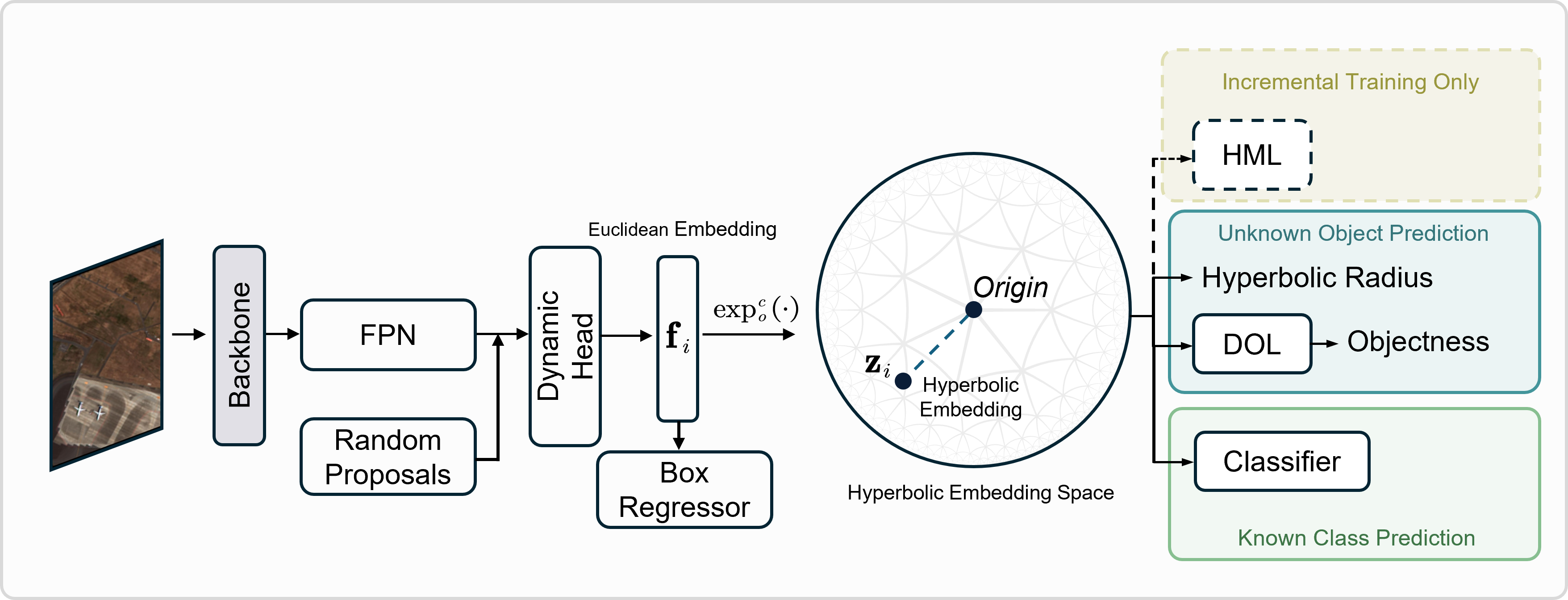}
    \caption{Overview of our proposed HyRS-OWOD framework. The backbone and FPN extract multi-scale features from an input image, which are processed together with randomly generated proposals 
    by the dynamic head to obtain proposal features $\mathbf{f}_i$. These features are used for bounding-box regression and mapped into the Poincar\'e ball via the exponential map $\exp_{\mathbf{0}}^{c}(\cdot)$. For unknown-object prediction, DOL estimates class-agnostic objectness, while HUL measures uncertainty using the hyperbolic radius; proposals with high objectness and high hyperbolic uncertainty are identified as unknown. In parallel, the classifier predicts known object categories. During incremental training, HML is additionally applied to learn discriminative representations and mitigate catastrophic forgetting.}
    \label{fig3}
\end{figure*}

The overall architecture of the proposed HyRS-OWOD framework is illustrated in Fig.~\ref{fig3}. HyRS-OWOD is built upon RandBox~\cite{32-wang2023random} and introduces three key components: (1) Decoupled Objectness Learning (DOL) (Sec.~\ref{sec:objectness}), which separates foreground from background; (2) Hyperbolic Uncertainty Learning (HUL) (Sec.~\ref{sec:uncertainty}), which distinguishes unknown objects from known classes; and (3) Hyperbolic Metric Learning (HML) (Sec.~\ref{sec:metric}), which supports learning novel classes while mitigating catastrophic forgetting.

Given an input image, the model first extracts feature maps and then randomly generates a set of candidate proposal boxes, which are fed into their exclusive dynamic head~\cite{65-sun2021sparse} to generate object features. The bounding box regression branch refines the proposal locations based on these object features. In parallel, the object features are projected into hyperbolic space to obtain hyperbolic embeddings. The hyperbolic embeddings are then fed into a known-class classification branch and a class-agnostic objectness branch. The objectness branch is implemented as a lightweight multilayer perceptron (MLP) that outputs an objectness score for each proposal. The hyperbolic embeddings are further used for uncertainty estimation and metric learning in the following modules. 

During training, we follow RandBox to employ the dynamic matcher~\cite{66-Chen2023DiffusionDet} to assign proposal boxes to ground-truth instances. The matched proposals are used for known class detection, while the remaining unmatched proposals with top objectness are selected as initial candidate unknowns and used to optimize HUL. At inference, unknown objects are directly identified based on their class-agnostic objectness and hyperbolic uncertainty. During incremental learning, we retain 50 exemplars from each previously learned class following the exemplar-replay protocols adopted in~\cite{3-joseph2021towards,31-Sun2024Orthogonality,5-Hu2026Unveiling}.

\subsection{Decoupled Objectness Learning}\label{sec:objectness}

Objectness should indicate whether a proposal corresponds to a foreground object, irrespective of its semantic category. To train the objectness branch, we adopt a binary cross-entropy loss:
\begin{equation}
\mathcal{L}_{\text{obj}} = - \sum_{b} \left[ y_b^{\text{fg}} \log(o_b) + (1 - y_b^{\text{fg}}) \log(1 - o_b) \right],
\end{equation}
where $o_b \in [0,1]$ and $y_b^{\text{fg}} \in \{0,1\}$ are the predicted objectness score and the ground-truth foreground/background label for proposal $b$, respectively.

Because the objectness branch is supervised using only known-class annotations, its predictions may become correlated with known-class confidence. Consequently, proposals containing unknown objects may receive low objectness scores and be incorrectly suppressed as background. To reduce this semantic bias, we introduce a decoupling loss that penalizes the linear correlation between the objectness scores and known-class predictions. Inspired by~\cite{31-Sun2024Orthogonality}, we measure their linear correlation using the squared correlation coefficient:
\begin{equation}
\mathcal{L}_{\text{dec}} = \sum_{c=1}^{C} \frac{\left( \mathrm{cov}(p_{b,c}, o_b) \right)^2}{\mathrm{var}(p_{b,c}) \, \mathrm{var}(o_b)},
\end{equation}
where covariance and variance are computed over all proposals within a mini-batch; $C$ denotes the number of known classes. By reducing the statistical dependence between objectness and known-class predictions, DOL encourages the objectness branch to capture class-agnostic foreground cues rather than category-specific semantic information.

\subsection{Hyperbolic Uncertainty Learning for Unknown Discrimination}\label{sec:uncertainty}

\begin{figure*}[t]
    \centering
    \includegraphics[width=0.98\textwidth]{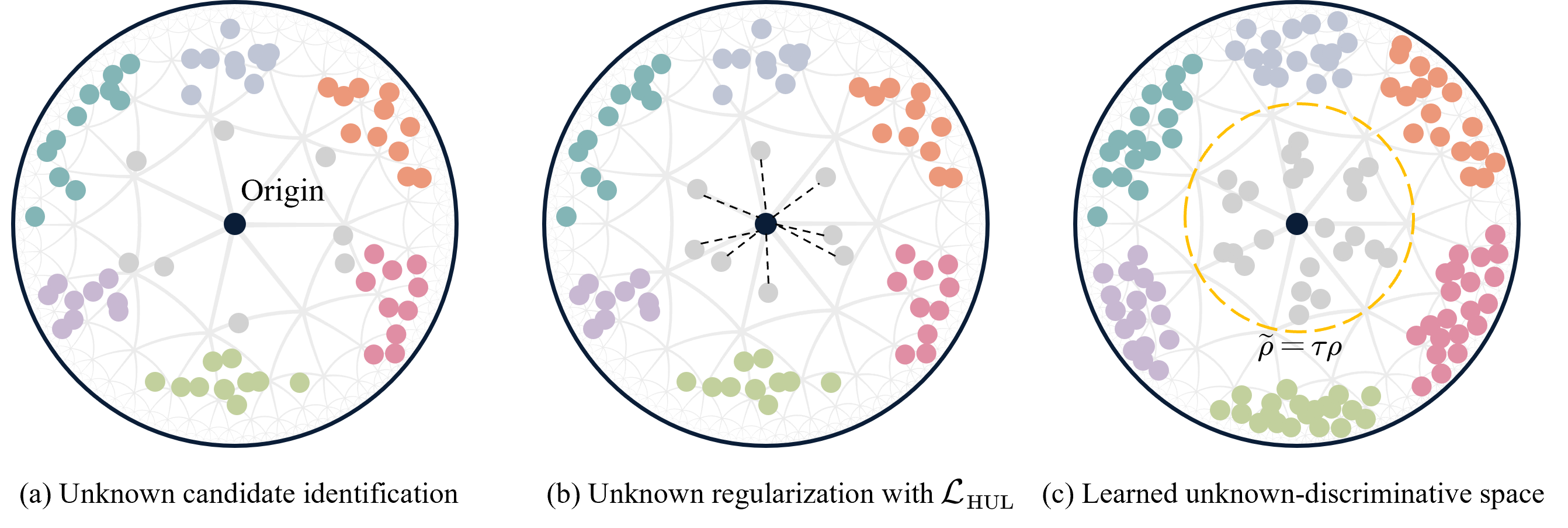}
    \caption{Illustration of the proposed HUL strategy for known--unknown discrimination. Different colors represent proposals from different known classes, gray points denote candidate unknown proposals. (a) Candidate unknown proposals are selected for uncertainty-aware learning. (b) The HUL loss $\mathcal{L}_{\mathrm{HUL}}$ encourages candidate unknown proposals to lie in the high-uncertainty region near the origin. (c) After training, uncertain unknown proposals tend to lie closer to the origin than confident known proposals. The yellow dashed circle represents the normalized radius threshold $\tilde{\rho}=\tau_{\rho}$ used for known--unknown discrimination.} 
    \label{fig2}
\end{figure*}

We estimate the uncertainty of each proposal by its radial position in the Poincar\'e ball. Given an input image $I$, let $\mathcal{P}_I = \{\mathbf{f}_i\}_{i=1}^{N_I} $ denote the set of proposal embeddings, where $\mathbf{f}_i \in \mathbb{R}^{D}$. Each proposal embedding is mapped into a $d$-dimensional Poincar\'e ball  $\mathbb{B}^{d}_{c} = \{\mathbf{z} \in \mathbb{R}^{d} : c \|\mathbf{z}\|_2^2 < 1\} $ with constant negative curvature $ -c $ by a lightweight hyperbolic projection head $g_{\phi}: \mathbb{R}^{D} \rightarrow \mathbb{R}^{d}$, followed by a projection onto the ball:
\begin{equation}
\mathbf{z}_i =
\Pi_{\mathbb{B}^{d}_{c}}\left(g_{\phi}(\mathbf{f}_i)\right)
=
\tanh\left(\sqrt{c} \, \|g_{\phi}(\mathbf{f}_i)\|_2\right)
\frac{g_{\phi}(\mathbf{f}_i)}
{\sqrt{c} \, \|g_{\phi}(\mathbf{f}_i)\|_2},
\end{equation}
where $ \mathbf{z}_i \in \mathbb{B}^{d}_{c} $ denotes the hyperbolic proposal embedding, and the zero vector case is defined by continuity.

we compute the hyperbolic radius of proposal $i$ as its distance to the origin~\cite{10-Khrulkov2020Hyperbolic,12-chen2023hyperbolic}:
\begin{equation}
    \rho_i = d_c(\mathbf{z}_i,\mathbf{0})
    =
    \frac{2}{\sqrt{c}}
    \operatorname{arctanh}
    \left(
    \sqrt{c}\|\mathbf{z}_i\|_2
    \right).
    \label{eq:hyperbolic_radius}
\end{equation}

Since the scale of the hyperbolic radius may vary across images, we normalize the radius within each image:
\begin{equation}
    \tilde{\rho}_i = \frac{\rho_i}{\max_{1 \le j \le N_I} \rho_j + \epsilon},
    \label{eq:normalized_radius}
\end{equation}
where \(\epsilon\) is a small constant to avoid division by zero. The hyperbolic uncertainty score is then defined as
\begin{equation}
    u_i = 1-\tilde{\rho}_i,
\label{eq:hyperbolic_uncertainty}
\end{equation}
where a larger $u_i$ indicates higher uncertainty. In this way, proposals close to the origin obtain larger uncertainty scores, whereas confident known proposals near the boundary receive smaller uncertainty scores.

To explicitly encourage unknown proposals to stay in the high-uncertainty region of the Poincar\'e ball, we introduce a smooth hyperbolic unknown loss to regularize candidate unknown proposals:
\begin{equation}
\mathcal{L}_{\mathrm{HUL}}
=\frac{1}{|\mathcal{U}|}\sum_{i\in\mathcal{U}}\frac{1}{\beta}\log\left(1+\exp\left(\beta(\tilde{\rho}_i-\tau_{\rho})\right)\right),
\label{eq:smooth_unknown_loss}
\end{equation}
where $\mathcal{U}$ denotes the set of candidate unknown proposals,
$\tau_{\rho}$ is the radius threshold, and $\beta$ controls the smoothness of the margin. When $\beta$ becomes large, Eq.~\eqref{eq:smooth_unknown_loss}
approaches the hard hinge loss $\max(0,\tilde{\rho}_i-\tau_{\rho})$. The overall HUL process is illustrated in Fig.~\ref{fig2}.

\subsection{Hyperbolic Metric Learning for Incremental Learning}
\label{sec:metric}
In this subsection, we introduce HML to mitigate catastrophic forgetting in incremental learning.  In particular, HML assigns larger weights to hard-negative proposal pairs that belong to different classes but are located close to each other.

At incremental step \(t\), let \(\mathcal{C}^{t}\) denote the set of all classes learned up to and including step \(t\). We maintain a class-balanced proposal buffer \(\mathcal{Q}\), which stores hyperbolic proposal embeddings from previously learned classes. Given the foreground proposal embeddings in the current mini-batch \(\mathcal{B}\), we construct the metric learning set as $\mathcal{A} = \mathcal{B} \cup \mathcal{Q}.$ For each anchor proposal embedding \(\mathbf{z}_{i} \in \mathcal{A}\) with class label \(y_{i} \in \mathcal{C}^{t}\), we define its positive and negative sets as
\begin{equation}
\left\{
\begin{aligned}
\mathcal{P}(i) &=\left\{p \in \mathcal{A} \setminus \{i\}:y_{p}=y_{i}\right\}, \\
\mathcal{N}(i) &=\left\{n \in \mathcal{A}:y_{n}\neq y_{i} \right\}.
\end{aligned}
\right.
\label{eq:positive_negative_sets}
\end{equation}
Only anchors with non-empty positive sets are used for metric learning, and the corresponding anchor index set is denoted as \(\mathcal{S}\).

In incremental learning, the most informative negative samples are typically those that lie close to the anchor in the embedding space but belong to different classes. To exploit this, we assign larger weights to hard negatives according to their Poincar\'e distances to the anchor:
\begin{equation}
\alpha_{i,n}^{-}
=\frac{\exp\left(
-d_{c}(\mathbf{z}_{i},\mathbf{z}_{n})/\tau_{\mathrm{h}}\right)}{\sum_{m \in \mathcal{N}(i)}
\exp\left(-d_{c}(\mathbf{z}_{i},\mathbf{z}_{m})/\tau_{\mathrm{h}}\right)},\quad n \in \mathcal{N}(i),
\label{eq:hard_negative_weight}
\end{equation}
where \(d_{c}(\cdot,\cdot)\) denotes the Poincar\'e distance and \(\tau_{\mathrm{h}}\) is a temperature parameter that controls the hardness distribution. A smaller distance yields a larger weight, thereby directing the model to focus on more confusing inter-class proposal pairs. We define the positive term and the hard-negative term as
\begin{equation}
\left\{
\begin{aligned}
A_{i}^{+}&=\sum_{p \in \mathcal{P}(i)}
\exp\left(-d_{c}(\mathbf{z}_{i},\mathbf{z}_{p})/\tau_{\mathrm{m}}\right), \\
A_{i}^{-}&=\sum_{n \in \mathcal{N}(i)}\alpha_{i,n}^{-}\exp\left(-d_{c}(\mathbf{z}_{i},\mathbf{z}_{n})/\tau_{\mathrm{m}}\right).
\end{aligned}
\right.
\label{eq:positive_negative_terms}
\end{equation}
where $\tau_{\mathrm{m}}$ is the metric learning temperature. 
HML loss is then written as
\begin{equation}
\mathcal{L}_{\mathrm{HML}}
=-\frac{1}{|\mathcal{S}|}\sum_{i \in \mathcal{S}}\log\frac{A_{i}^{+}}{A_{i}^{+}+A_{i}^{-}}.
\label{eq:weighted_hml_loss}
\end{equation}
This objective pulls same-class proposal embeddings closer together in the Poincar\'e ball while pushing hard negatives from different classes apart. By emphasizing the most confusing inter-class pairs, the model learns more discriminative class boundaries during incremental learning.

The proposal buffer \(\mathcal{Q}\) is updated after each mini-batch by storing foreground proposal embeddings along with their corresponding labels. To prevent class imbalance, we retain at most \(K_{\max}\) embeddings per class and update the buffer in a first-in-first-out manner. 

\subsection{Overall Training Objective}
\label{sec:objective}

The detection loss for known objects consists of a classification loss $\mathcal{L}_{cls}$ and a bounding box regression loss $\mathcal{L}_{reg}$: 
\begin{equation}
\mathcal{L}_{\mathrm{kn}}=\mathcal{L}_{cls}+\mathcal{L}_{reg},
\end{equation}
where $\mathcal{L}_{cls}$ is the focal loss~\cite{67-lin2017focal} for class imbalance and the $\mathcal{L}_{reg}$ is smooth ${L}_{1}$ loss~\cite{68-girshick2015fast}.

The overall training objective is therefore 
\begin{equation}
\mathcal{L}_{\mathrm{base}}
=
\mathcal{L}_{\mathrm{kn}}
+
\lambda_{\mathrm{obj}}\mathcal{L}_{\mathrm{obj}}
+
\lambda_{\mathrm{dec}}\mathcal{L}_{\mathrm{dec}}
+
\lambda_{\mathrm{HUL}}\mathcal{L}_{\mathrm{HUL}},
\label{eq:base_objective}
\end{equation}
where $\lambda_{\mathrm{obj}}$, $\lambda_{\mathrm{dec}}$, and $\lambda_{\mathrm{HUL}}$ are the corresponding loss weights.

In the incremental learning stage, we further introduce the weighted HML loss $\mathcal{L}_{\mathrm{HML}}$. The overall incremental training objective is defined as
\begin{equation}
\mathcal{L}_{\mathrm{inc}} = \mathcal{L}_{\mathrm{base}} + \lambda_{\mathrm{HML}}\mathcal{L}_{\mathrm{HML}},
\label{eq:incremental_objective}
\end{equation}
where $\lambda_{\mathrm{HML}}$ controls the contribution of the weighted HML objective.

\section{Experiments}
In this section, we first describe the datasets, evaluation metrics, and implementation details. We then present quantitative and qualitative comparisons against baseline and state-of-the-art approaches. Finally, we conduct ablation studies to validate the effectiveness of the proposed components.

\subsection{Datasets}
\subsubsection{NWPU VHR-10}
This dataset consists of 650 positive images, categorized into ten geospatial object classes for object detection, including airplane (AP), vehicle (VH), ship (SH), harbor (HB), baseball diamond (BD), basketball court (BC), tennis court (TC), ground track field (GTF), storage tank (ST), and bridge (BR). The spatial resolution of the images ranges from 0.5 to 2 m.


\subsubsection{DIOR}
This dataset comprises 23,463 aerial images, containing 192,472 object instances across 20 categories: airplane (AP), windmill (WM), airport (AT), train station (TS), baseball field (BF), tennis court (TC), bridge (BR), storage tank (ST), chimney (CM), ship (SH), dam (DM), expressway service area (EA), stadium (SD), expressway toll station (ES), harbor (HB), golf course (GF), vehicle (VH), ground track field (GTF), overpass (OP), and basketball court (BC). The spatial resolution ranges from 0.5 to 30 m, and each image has a fixed size of $800 \times 800$ pixels.


\subsubsection{DOTA-v1.5}
This dataset comprises 2,806 large-scale aerial images, with dimensions ranging from $800 \times 800$ to $4,000 \times 4,000$  pixels. It contains 402,089 annotated instances across 16 categories: basketball court, baseball diamond, bridge, container crane, ground track field, harbor, helicopter, large vehicle, plane, roundabout, small vehicle, ship, storage tank, soccer ball field, swimming pool, and tennis court.
 
\subsubsection{Dataset Splits for OWOD}
\label{sec:split}
For a fair comparison, we adopt the same known--unknown class partitions as in~\cite{5-Hu2026Unveiling}, including the 16+4, 10+10, and 4+16 splits on DIOR, the 8+8 split on DOTA, and the 8+2 split on NWPU VHR-10. In addition to these benchmark settings, we establish a progressive OWOD protocol on NWPU VHR-10 to further evaluate the model under sequential category discovery, as summarized in Table~\ref{tab3}. In Task~1, seven categories (AP, BC, BD, BR, GTF, HB, and SH) are used as the known classes, whereas ST, TC, and VH are regarded as unknown classes. ST, TC, and VH are then sequentially introduced as newly annotated classes in Tasks~2, 3, and 4, respectively. This protocol evaluates the model's ability to discover unknown objects, incrementally learn novel categories, and retain knowledge of previously learned classes. Furthermore, to systematically evaluate the model's performance, we conduct additional experiments on DIOR under the 15+5 and 19+1 settings following the protocols commonly adopted in generic-domain OWOD methods~\cite{3-joseph2021towards,30-Zohar2023PROB,31-Sun2024Orthogonality,32-wang2023random}. For clarity, the novel classes are highlighted with a gray background in Table~\ref{tab4}.

\subsection{Evaluation Metrics}
We follow the evaluation protocol established in~\cite{3-joseph2021towards,30-Zohar2023PROB,31-Sun2024Orthogonality}. For known classes, we report mean average precision (mAP) to evaluate the overall detection performance. In the incremental learning setting, mAP is reported separately for known classes and novel classes. For unknown classes, we adopt unknown recall (U-Recall) as the primary metric for evaluating the model's ability to retrieve unknown objects. We further report wilderness impact (WI)~\cite{59-DhamijaOverlooked2020}, which measures the influence of unknown objects on the precision of known class detection, and absolute open-set error (A-OSE)~\cite{60-Miller2018Dropout}, which counts the number of unknown objects misclassified as known classes. 

\subsection{Implementation Details} 
We build our model upon RandBox~\cite{32-wang2023random}, using ResNet-50~\cite{61-He2016Deep} with a Feature Pyramid Network (FPN)~\cite{62-lin2017feature} as the backbone. All experiments are conducted on a single NVIDIA GeForce RTX 5090 GPU with a batch size of 8. The model is optimized using AdamW with a base learning rate of $3\times10^{-5}$ and a weight decay of $1\times10^{-4}$. A warmup schedule is applied over the first 33 iterations. The learning rate starts at $3\times10^{-7}$, corresponding to a warmup factor of 0.01, and linearly increases to the base learning rate of $3\times10^{-5}$. The base training stage is conducted for 30 epochs on NWPU VHR-10, DIOR, and DOTA. Each incremental learning task is also trained for 30 epochs.

\begin{table*}[!t]
\centering
\caption{Open-World Object Detection performance on DIOR (Top), DOTA (Middle), and NWPU VHR-10 (Bottom) under various settings. The best results within each setting are highlighted in \textbf{bold}.}
\label{tab2}
\setlength{\tabcolsep}{6.5pt} 
\small
\begin{tabular}{l|cc|ccc}
\cline{1-6} 
\multicolumn{1}{c|}{\multirow{3}*{\textbf{Task IDs ($\rightarrow$)}}} & \multicolumn{2}{c|}{\textbf{Task 1 (Base Stage)}} & \multicolumn{3}{c}{\textbf{Task 2 (Incremental Stage)}} \\
\cline{2-6} 
\multicolumn{1}{c|}{} & \multirow{2}{*}{U-Recall ($\uparrow$)} & \multirow{2}{*}{mAP ($\uparrow$)} & \multicolumn{3}{c}{mAP ($\uparrow$)} \\
\cline{4-6}
\multicolumn{1}{c|}{} & & & Previously known & Currently known & Both \\
\midrule
\multicolumn{6}{l}{\hspace*{0.5em}\textbf{DIOR (16 + 4 Setting)}} \\
\midrule
ORE~\cite{3-joseph2021towards}& 14.18 & 50.62 & 46.30 & 67.11 & 50.46 \\
OW-DETR~\cite{27-Gupta2022OW-DETR}& 18.53 & 51.58 & 49.71 & 67.46 & 53.26 \\
PROB~\cite{30-Zohar2023PROB}& 23.37 & 53.90 & 51.61 & 71.01 & 55.49 \\
KTCN~\cite{63-Xi2024KTCN}& 27.59 & 55.88 & 52.96 & 70.90 & 56.55 \\
SGROD~\cite{64-He2025Recalling}& 35.82 & 55.68 & 52.22 & 71.22 & 56.02 \\
Hu et al.~\cite{5-Hu2026Unveiling}& 42.93 & 57.69 & 54.23 &  \textbf{75.47} &  58.48 \\
\hline 
Base model &39.07 & 69.15 & 59.51 & 63.74 & 60.36 \\
\textbf{Ours} &  \textbf{52.93} &  \textbf{71.26}&  \textbf{68.59} & 72.57&  \textbf{69.39}\\
\midrule
\midrule
\multicolumn{6}{l}{\hspace*{0.5em}\textbf{DIOR (10 + 10 Setting)}} \\
\midrule
ORE~\cite{3-joseph2021towards}& 21.41 & 54.62 & 49.61 & 53.11 & 51.36 \\
OW-DETR~\cite{27-Gupta2022OW-DETR}& 28.01 & 56.58 & 49.54 & 55.20 & 52.37 \\
PROB~\cite{30-Zohar2023PROB}& 27.77 & 57.90 & 54.21 & 55.01 & 54.61 \\
KTCN~\cite{63-Xi2024KTCN}& 33.10 & 59.22 & 56.12 & 58.12 & 57.12 \\
SGROD~\cite{64-He2025Recalling}& 39.25 & 59.17 & 55.02 & 61.11 & 58.07 \\
Hu et al.~\cite{5-Hu2026Unveiling}& 46.33 & 62.51 & 57.09 & 63.21 & 60.15 \\
\hline 
Base model & 34.56 & 71.13 & 58.31 & 60.69 & 59.50\\
\textbf{Ours} & \textbf{53.72}& \textbf{73.49}& \textbf{61.14}& \textbf{65.97} & \textbf{63.56}\\
\midrule
\midrule
\multicolumn{6}{l}{\hspace*{0.5em}\textbf{DIOR (4 + 16 Setting)}} \\
\midrule
ORE~\cite{3-joseph2021towards}& 16.77 & 58.90 & 59.04 & 53.66 & 54.74 \\
OW-DETR~\cite{27-Gupta2022OW-DETR}& 21.08 & 64.62 & 62.44 & 54.60 & 56.17 \\
PROB~\cite{30-Zohar2023PROB}& 19.57 & 67.10 & 61.91 & 51.17 & 53.32 \\
KTCN~\cite{63-Xi2024KTCN}& 23.67 & 65.34 & 59.76 & 57.89 & 58.26 \\
SGROD~\cite{64-He2025Recalling} & 31.77 & 69.18 & 63.17 & 56.92 & 58.17 \\
Hu et al.~\cite{5-Hu2026Unveiling}& \textbf{41.61} & 68.79 & 67.39 & 59.26 & 60.89 \\
\hline
Base model & 23.21 & 73.10& 63.50&58.83 & 59.76\\
\textbf{Ours} &40.11 &\textbf{79.48} &\textbf{70.84} & \textbf{60.98}& \textbf{62.96}\\
\midrule
\midrule
\multicolumn{6}{l}{\hspace*{0.5em}\textbf{DOTA (8 + 8 Setting)}} \\
\midrule
ORE~\cite{3-joseph2021towards}& 15.01 & 41.92 & 39.72 & 56.11 & 47.92 \\
OW-DETR~\cite{27-Gupta2022OW-DETR}& 13.79 & 43.55 & 39.23 & 57.01 & 48.12 \\
PROB~\cite{30-Zohar2023PROB}& 19.05 & 43.09 & 40.06 & 59.01 & 49.54 \\
KTCN~\cite{63-Xi2024KTCN}& 25.02 & 46.80 & 44.99 & 58.60 & 51.80 \\
SGROD~\cite{64-He2025Recalling}& 31.90 & 45.11 & 43.69 & 59.96 & 51.82 \\
Hu et al.~\cite{5-Hu2026Unveiling} & 36.44 & 47.99 & 45.84 & \textbf{61.24} & \textbf{53.54} \\
\hline
Base model & 28.35&45.53 & 41.63&49.99 & 45.81\\
\textbf{Ours} & \textbf{36.85}&\textbf{49.36} & \textbf{47.72}& 58.47& 53.10\\
\midrule 
\midrule
\multicolumn{6}{l}{\hspace*{0.5em}\textbf{NWPU VHR-10 (8 + 2 Setting)}} \\
\midrule
ORE~\cite{3-joseph2021towards}& 27.18 & 83.19 & 81.30 & 81.40 & 81.32 \\
OW-DETR~\cite{27-Gupta2022OW-DETR}& 26.52 & 83.58 & 82.15 & 78.70 & 81.46 \\
PROB~\cite{30-Zohar2023PROB}& 38.49 & 84.16 & 84.21 & 78.01 & 82.97 \\
KTCN~\cite{63-Xi2024KTCN}& 36.74 & 83.78 &86.50 & 80.90 & 85.38 \\
SGROD~\cite{64-He2025Recalling}& 46.82 & 82.68 & 83.72 & 81.02 & 83.18 \\
Hu et al.~\cite{5-Hu2026Unveiling} & 55.90 & 86.03 & 86.23 & \textbf{85.73} & 86.13 \\
\hline
Base model & 45.18&\textbf{91.94} & 91.47&76.85 &88.54 \\
\textbf{Ours} &\textbf{58.94} &89.06 &\textbf{91.57} & 84.86&\textbf{90.23} \\
\bottomrule
\end{tabular}%

\end{table*}

\begin{table*}[!t]
\centering
\caption{Evaluation of Unknown Object confusion on NWPU VHR-10.}
\small

\resizebox{\textwidth}{!}{
\begin{tabular}{l | c c c c | c c c c | c c c c | c}
\toprule
\label{tab3}
\textbf{Task IDs ($\rightarrow$)}
& \multicolumn{4}{c|}{\textbf{Task 1}}
& \multicolumn{4}{c|}{\textbf{Task 2}}
& \multicolumn{4}{c|}{\textbf{Task 3}}
& \textbf{Task 4} \\
\midrule

\multirow{2}{*}{Method}
& \multirow{2}{*}{mAP}& U-Recall & WI & A-OSE
& \multirow{2}{*}{mAP}& U-Recall & WI & A-OSE
& \multirow{2}{*}{mAP}& U-Recall & WI & A-OSE
 & mAP \\

&& \multirow{-1}{*}{$(\uparrow)$}
& \multirow{-1}{*}{$(\downarrow)$}
& \multirow{-1}{*}{$(\downarrow)$}
&& \multirow{-1}{*}{$(\uparrow)$}
& \multirow{-1}{*}{$(\downarrow)$}
& \multirow{-1}{*}{$(\downarrow)$}
&& \multirow{-1}{*}{$(\uparrow)$}
& \multirow{-1}{*}{$(\downarrow)$}
& \multirow{-1}{*}{$(\downarrow)$}
& \multirow{-1}{*}{$(\uparrow)$} \\

\midrule

ORE~\cite{3-joseph2021towards}
&88.97 	&27.33 	&0.000000 	&2 	&89.74 	&30.18 	&0.004090 	&3 	&87.40 	&34.23 	&0.006612 	&15 	&87.22 
 \\

OW-DETR~\cite{27-Gupta2022OW-DETR}
&85.95 	&24.07 	&0.000000 	&9 	&85.39 	&25.76 	&0.000129 	&8 	&85.60 	&27.34 	&0.001691 	&7 	&82.98 
 \\

PROB~\cite{30-Zohar2023PROB}
&86.60 	&36.21 	&0.000000 	&0 	&82.49 	&31.33 	&0.000000 	&0 	&78.24 	&29.68 	&0.000000 	&0 	&84.88 
 \\
KTCN~\cite{63-Xi2024KTCN} 
&82.62 	&33.34 	&0.000234 	&12 	&78.27 	&37.94 	&0.000311 	&15 	&77.16 	&36.06 	&0.000707 	&3 	&82.48 
\\

SGROD~\cite{64-He2025Recalling}
&80.48 	&35.25 	&0.000000 	&0 	&84.24 	&40.46 	&0.000000 	&0 	&81.33 	&39.35 	&0.000000 	&0 	&83.90 
\\
\midrule

Base model
&90.48 	&38.39 	&0.000000 	&1 	&83.76 	&45.54 	&0.000000 	&13 	&84.56 	&51.97 	&0.000000 	&2 	&84.04 
\\
 
\textbf{Ours}
& \textbf{93.74 }	&\textbf{55.28} 	&0.000578 	&2 	&\textbf{94.46} 	&\textbf{60.57} 	&0.000214 	&2 	&\textbf{93.63} 	&\textbf{63.01} 	&0.000000 	&0 	&\textbf{92.39} 
\\
\bottomrule
\end{tabular}
}
\end{table*}

\subsection{Quantitative Results}
 
 \textit{Open-world object detection.} 
 We compare our method with other state-of-the-art methods on DIOR, DOTA, and NWPU VHR-10 datasets under different incremental settings and summarize the results in Table~\ref{tab2}. Following the experimental protocol of Hu et al.~\cite{5-Hu2026Unveiling}, we report U-Recall and mAP after base training, together with the mAP of previously known, currently known, and overall performance after incremental learning. For unknown object detection, our method achieves the best performance in four of the five settings, surpassing the previous best results by 10.00, 7.39, 0.41, and 3.04 percentage points in DIOR 16+4, DIOR 10+10, DOTA 8+8, and NWPU VHR-10 8+2, respectively. In terms of known object detection, our method outperforms the competing methods in most settings, achieving the highest mAP after base training in four of the five settings. After incremental learning, our method achieves the best performance on previously known classes across all five settings, with improvements of 14.36, 4.05, 3.45, 1.88, and 5.07 percentage points, respectively. Moreover, our method obtains the highest overall mAP in four of the five settings, outperforming the previous best results by 10.91, 3.41, 2.07, and 4.10 percentage points in DIOR 16+4, DIOR 10+10, DIOR 4+16, and NWPU VHR-10 8+2, respectively, further demonstrating the superiority of our method in maintaining previously learned knowledge while achieving strong overall detection performance after incremental learning.
 
To further evaluate the confusion between unknown and known objects, we conduct additional experiments on NWPU VHR-10 using WI and A-OSE as evaluation metrics, with the results reported in Table~\ref{tab3}. Our method consistently achieves the highest U-Recall across all three open-world stages, outperforming the best competing methods by 19.07, 20.11, and 23.66 percentage points in Tasks 1–3, respectively. Meanwhile, the WI remains at a very low level, with values of only 0.000578 and 0.000214 in Tasks 1 and 2, and further decreases to zero in Task 3. A similar trend can be observed for A-OSE, where only two unknown objects are misclassified as known classes in Tasks 1 and 2, while no such confusion occurs in Task 3. These results show that the substantial gains in unknown-object recall are achieved while maintaining low absolute WI and A-OSE values.

\textit{Incremental object detection.}
To further evaluate the class-incremental detection performance of our method, we compare it with existing OWOD approaches following the evaluation protocols adopted in~\cite{3-joseph2021towards,27-Gupta2022OW-DETR,30-Zohar2023PROB}. Table~\ref{tab4} reports the results on DIOR under two class-incremental settings. Although the RandBox baseline already achieves competitive performance, our method further improves the overall mAP by 1.7 and 4.9 percentage points in the two settings, respectively. These improvements demonstrate the effectiveness of the proposed method in learning novel classes while maintaining the performance of previously learned classes.

\begin{table*}[!t]
\centering
\caption{Incremental object detection results on DIOR.}
\label{tab4}
\scriptsize
\setlength{\tabcolsep}{3.2pt} 
\renewcommand{\arraystretch}{1.25} 
\begin{tabular}{l cccccccccccccccccccccc}
\toprule
\textbf{15 + 5 setting} & AP & AT & BF & BC & BR & CM & DM & EA & ES & GF & GTF & HB & OP & SH & SD & ST & TC & TS & VH & WM & mAP \\ 
\midrule
ORE~\cite{3-joseph2021towards} & 55.5	&64.7	&64.6	&83.7	&24.8	&75.8	&42.6	&53.1	&43.6	&71.9	&64.4	&46.9	&45.5	&35.0	&53.5	&\cellcolor{gray!20}43.7	&\cellcolor{gray!20}77.9	&\cellcolor{gray!20}35.9	&\cellcolor{gray!20}52.0	&\cellcolor{gray!20}57.4	&54.6 \\
OW-DETR~\cite{27-Gupta2022OW-DETR} & 71.6	&59.6	&53.1	&56.4	&26.8	&71.6	&46.1	&44.0	&38.7	&73.3	&41.9	&36.6	&39.6	&32.1	&58.3	&\cellcolor{gray!20}53.7	&\cellcolor{gray!20}78.4	&\cellcolor{gray!20}47.4	&\cellcolor{gray!20}41.3	&\cellcolor{gray!20}74.1	&52.2 \\
PROB~\cite{30-Zohar2023PROB} &81.7	&63.2	&83.2	&61.8	&28.0	&81.4	&48.1	&45.4	&49.5	&66.1	&56.9	&37.0	&42.6	&28.0	&71.1	&\cellcolor{gray!20}63.0	&\cellcolor{gray!20}74.4	&\cellcolor{gray!20}38.9	&\cellcolor{gray!20}43.0	&\cellcolor{gray!20}47.4	&55.5 \\

KTCN~\cite{63-Xi2024KTCN}
&72.5	&56.4	&70.2	&80.9	&27.2	&76.1	&45.3	&51.8	&50.6	&70.5	&66.1	&44.8	&45.8	&70.7	&50.7	&\cellcolor{gray!20}60.3	&\cellcolor{gray!20}82.3	&\cellcolor{gray!20}25.3	&\cellcolor{gray!20}46.8	&\cellcolor{gray!20}68.8	&58.1
\\
SGROD~\cite{64-He2025Recalling}
&85.9	&64.9	&85.6	&55.4	&14.1	&87.3	&51.4	&33.1	&39.0	&70.6	&59.3	&3.7	&47.7	&30.9	&66.4	&\cellcolor{gray!20}76.1	&\cellcolor{gray!20}85.5	&\cellcolor{gray!20}46.1	&\cellcolor{gray!20}55.2	&\cellcolor{gray!20}67.4	&56.3
\\
\midrule
Base model &89.5	&61.9	&83.7	&67.3	&27.0	&86.2	&47.2	&45.8	&51.3	&65.7	&58.5	&44.4	&47.7	&11.2	&74.6	&\cellcolor{gray!20}79.2	&\cellcolor{gray!20}87.4	&\cellcolor{gray!20}42.9	&\cellcolor{gray!20}70.3	&\cellcolor{gray!20}67.8	&60.5\\

\textbf{Ours}&89.5	&68.9	&88.4	&71.0	&29.3	&86.4	&50.4	&58.1	&53.9	&72.1	&57.3	&40.9	&44.3	&22.1	&74.1	&\cellcolor{gray!20}76.7	&\cellcolor{gray!20}86.6	&\cellcolor{gray!20}37.1	&\cellcolor{gray!20}66.4	&\cellcolor{gray!20}70.4	&\textbf{62.2} \\ 
\midrule
\midrule
\textbf{19 + 1 setting} & AP & AT & BF & BC & BR & CM & DM & EA & ES & GF & GTF & HB & OP & SH & SD & ST & TC & TS & VH & WM & mAP \\ 
\midrule
ORE~\cite{3-joseph2021towards} & 55.1 & 63.4 & 67.2 & 86.2 & 29.0 & 79.2 & 41.8 & 53.1 & 46.9 & 68.4 & 66.5 & 49.3 & 50.6 & 67.3 & 68.9 & 55.5 & 80.0 & 45.1 & 41.1 & \cellcolor{gray!20}59.8 & 60.7 \\
OW-DETR~\cite{27-Gupta2022OW-DETR} & 64.0 & 60.5 & 65.9 & 78.9 & 31.2 & 73.9 & 41.8 & 50.6 & 44.3 & 67.2 & 48.9 & 28.3 & 48.4 & 68.7 & 29.4 & 49.8 & 76.4 & 44.2 & 38.4 & \cellcolor{gray!20}73.8 & 54.2 \\
PROB~\cite{30-Zohar2023PROB}& 83.2 & 55.2 & 86.3 & 70.5 & 28.2 & 81.3 & 34.8 & 43.6 & 52.7 & 51.6 & 55.9 & 12.7 & 41.3 & 52.1 & 64.9 & 72.2 & 82.9 & 40.1 & 49.1 & \cellcolor{gray!20}45.0 & 55.2 \\
KTCN~\cite{63-Xi2024KTCN}
&77.7	&57.5	&70.8	&85.5	&28.5	&75.0	&45.6	&52.9	&52.9	&67.9	&66.2	&47.5	&49.6	&70.2	&62.1	&60.1	&80.9	&33.3	&39.1	&\cellcolor{gray!20}68.0	&59.6
\\

SGROD~\cite{64-He2025Recalling}
&89.4	&51.0	&90.4	&74.6	&22.6	&87.9	&38.7	&59.0	&53.2	&61.3	&56.8	&13.2	&48.0	&29.8	&71.3	&75.3	&84.2	&42.0	&51.6	&\cellcolor{gray!20}63.0	&58.2
\\
\midrule
Base model &79.2	&66.6	&87.6	&78.0	&35.7	&88.2	&51.0	&62.7	&55.5	&76.1	&56.8	&55.4	&55.0	&11.5	&86.1	&75.3	&85.8	&47.4	&45.5	&\cellcolor{gray!20}76.6	&63.8
 \\ 
\textbf{Ours}  &90.5	&75.4	&92.5	&79.8	&39.0	&84.4	&50.6	&65.8	&56.8	&78.5	&68.7	&46.7	&57.8	&68.1	&77.4	&78.5	&88.5	&39.1	&58.8	&\cellcolor{gray!20}76.9	&\textbf{68.7} \\ 

\bottomrule
\end{tabular}
\end{table*}

\subsection{Qualitative Results}
\begin{figure*}[htbp]
    \centering
    \includegraphics[width=0.98\textwidth]{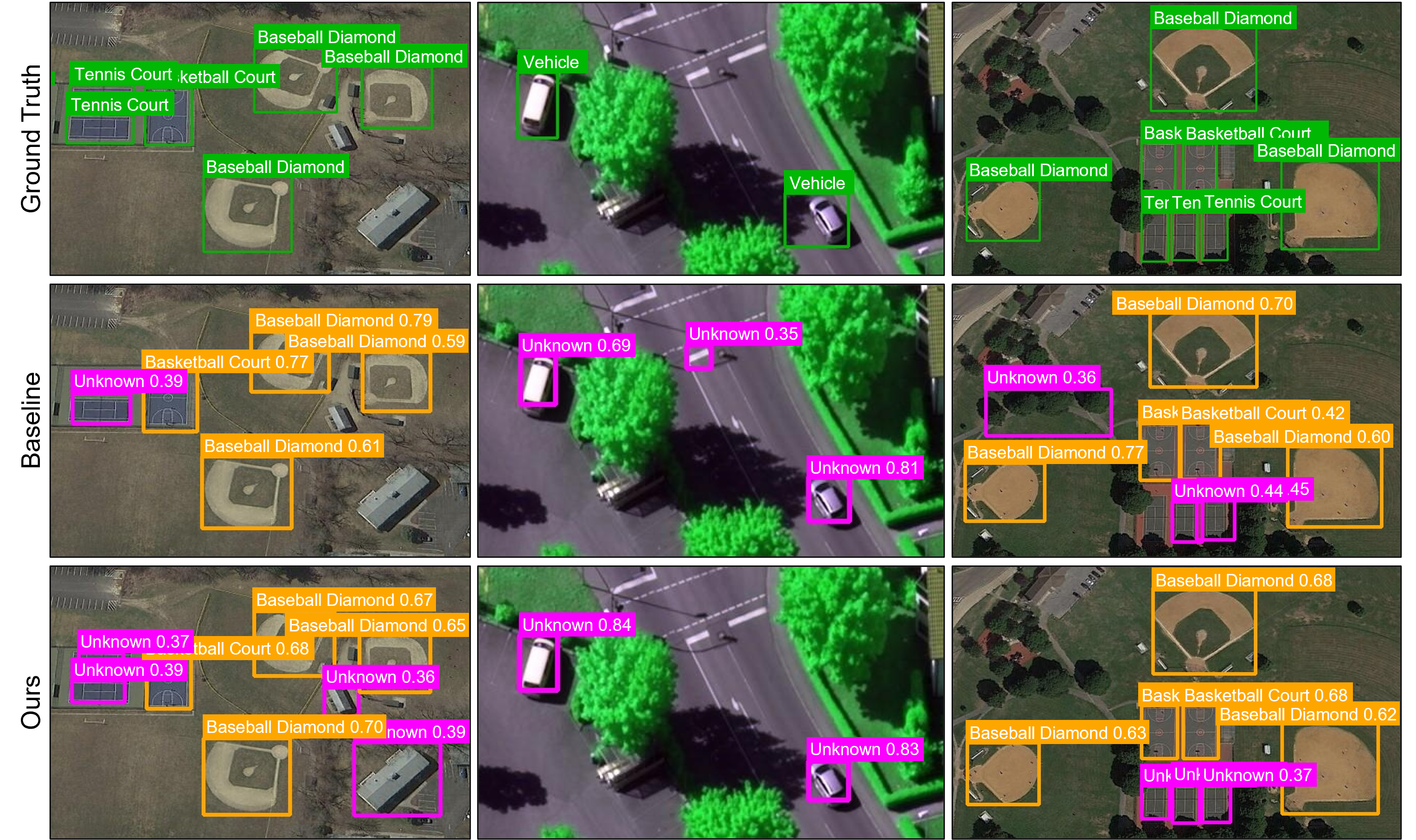}
    \caption{Qualitative comparison on the NWPU VHR-10 dataset. The first row shows the ground-truth annotations (\textcolor{green}{green}), while the second and third rows show the predictions of the baseline and our method, respectively. \textcolor{Orange}{Orange} and \textcolor{magenta}{magenta} boxes denote known and unknown objects, respectively.}
    \label{fig4}
\end{figure*}

\begin{figure*}[htbp]
    \centering
    \small
    \includegraphics[width=0.98\textwidth]{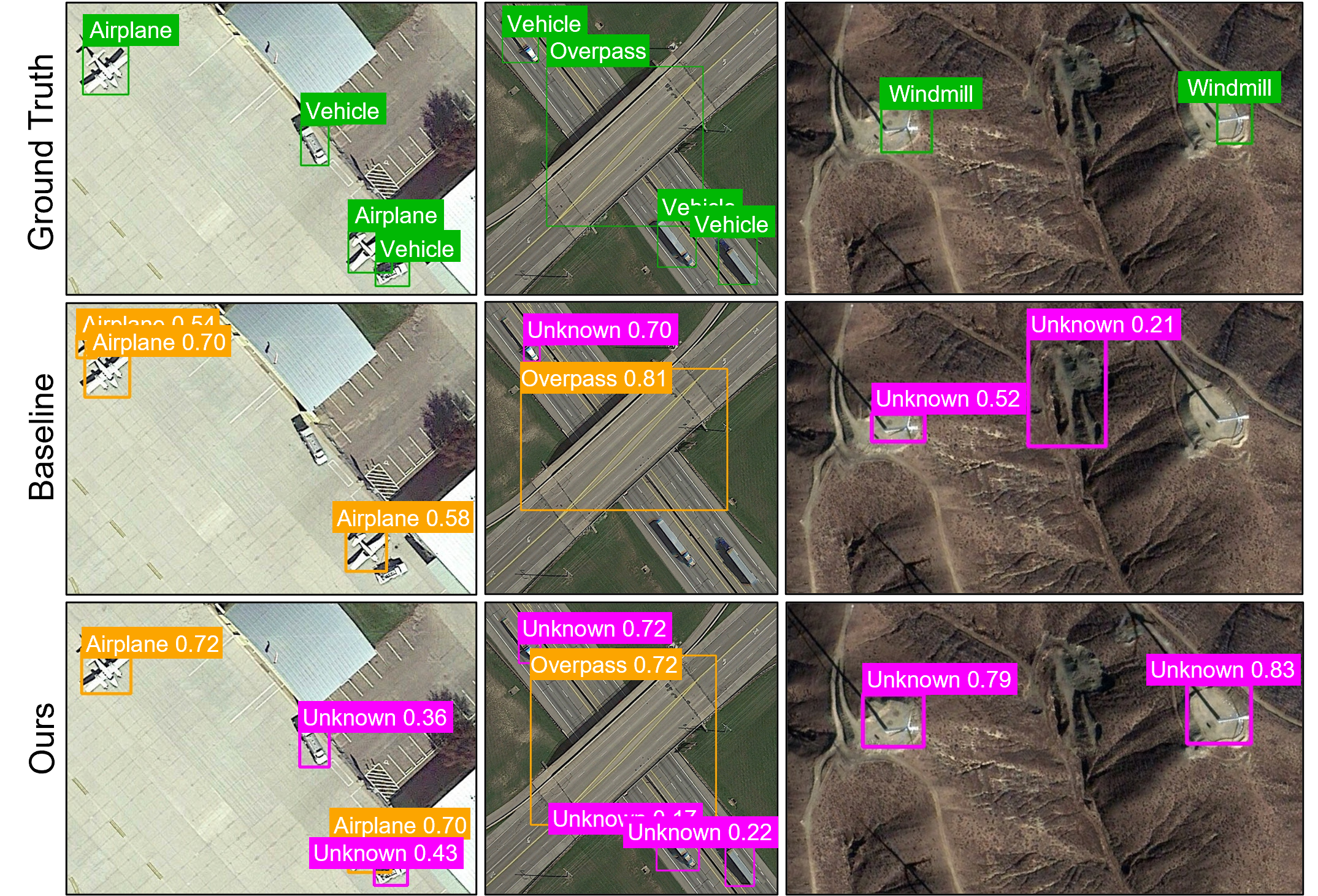}
    \caption{Qualitative comparison on the DIOR dataset. The first row shows the ground-truth annotations (\textcolor{green}{green}), while the second and third rows show the predictions of the baseline and our method, respectively. \textcolor{Orange}{Orange} and \textcolor{magenta}{magenta} boxes denote known and unknown objects, respectively.}
    \label{fig5}
\end{figure*}

Fig.~\ref{fig4} presents qualitative comparisons between our method and the RandBox baseline on NWPU VHR-10. Compared with the baseline, our method discovers more unknown objects while maintaining accurate detection of known classes. In the left example, our method identifies more tennis courts as unknown while correctly detecting the known basketball court and baseball diamonds, demonstrating improved unknown-object recall without compromising known-class detection. In the middle example, both vehicles are correctly identified as unknown with higher confidence scores, while the false-positive prediction produced by the baseline is suppressed. Similarly, in the right example, our method identifies individual tennis courts as unknown with more accurate localization while preserving reliable predictions for the known baseball diamonds and basketball courts. These qualitative observations are in line with the improvements in U-Recall, WI, and A-OSE reported in Table~\ref{tab3}, showing that our method improves unknown-object discovery while reducing the misclassification of unknown objects as known classes. Similar improvements are observed on DIOR, as illustrated in Fig.~\ref{fig5}. Our method detects more unknown vehicles and windmills that are missed by the baseline while maintaining reliable detection of known objects and suppressing false-positive unknown predictions. These results further demonstrate the effectiveness of our method across different remote sensing datasets and object categories.


\subsection{Ablation Study}
\label{sec:ablation}

\begin{table}[htbp]
\centering
\caption{Ablation study of the proposed components on NWPU VHR-10 under the 8+2 setting.}
\normalsize
\label{tab5}
\setlength{\tabcolsep}{3.2pt} 
\renewcommand{\arraystretch}{1.25} 
\resizebox{\linewidth}{!}{%
\begin{tabular}{l c c| c c c c}
\toprule
\multirow{2}{*}{\textbf{Task 2}} &\multirow{2}{*}{DOL}  & \multirow{2}{*}{HUL} & U-Recall & K-mAP& WI  & A-OSE\\
& &  &($\uparrow$) &($\uparrow$)&($\downarrow$) & ($\downarrow$)\\
\midrule
Base model & – & –&45.2	&91.9	&0.0002	&8 \\
&\CheckmarkBold  &\XSolidBrush &48.1	&90.2	&0.0001	&10 \\
&\XSolidBrush &\CheckmarkBold  & 51.2	&89	&0	&0
 \\
\textbf{Ours} &\CheckmarkBold &\CheckmarkBold &58.9	&89.1	&0.0000	&1 \\
\bottomrule
\end{tabular}%
}

\vspace{1em} 
\resizebox{\linewidth}{!}{%
\begin{tabular}{l  c| c c c}
\toprule
\multirow{2}{*}{\textbf{Task 2}} &\multirow{2}{*}{HML} & Previously & Currently & \\
 &   & known &  known &\multirow{-2}{*}{Both} \\
\midrule
Base model & – & 91.5	& 76.9	& 88.5 \\
Ours &\CheckmarkBold &91.6	&91.2	&91.5\\
\bottomrule
\end{tabular}%
}

\end{table}

\begin{figure}[htbp]
    \centering
    \includegraphics[width=\columnwidth]{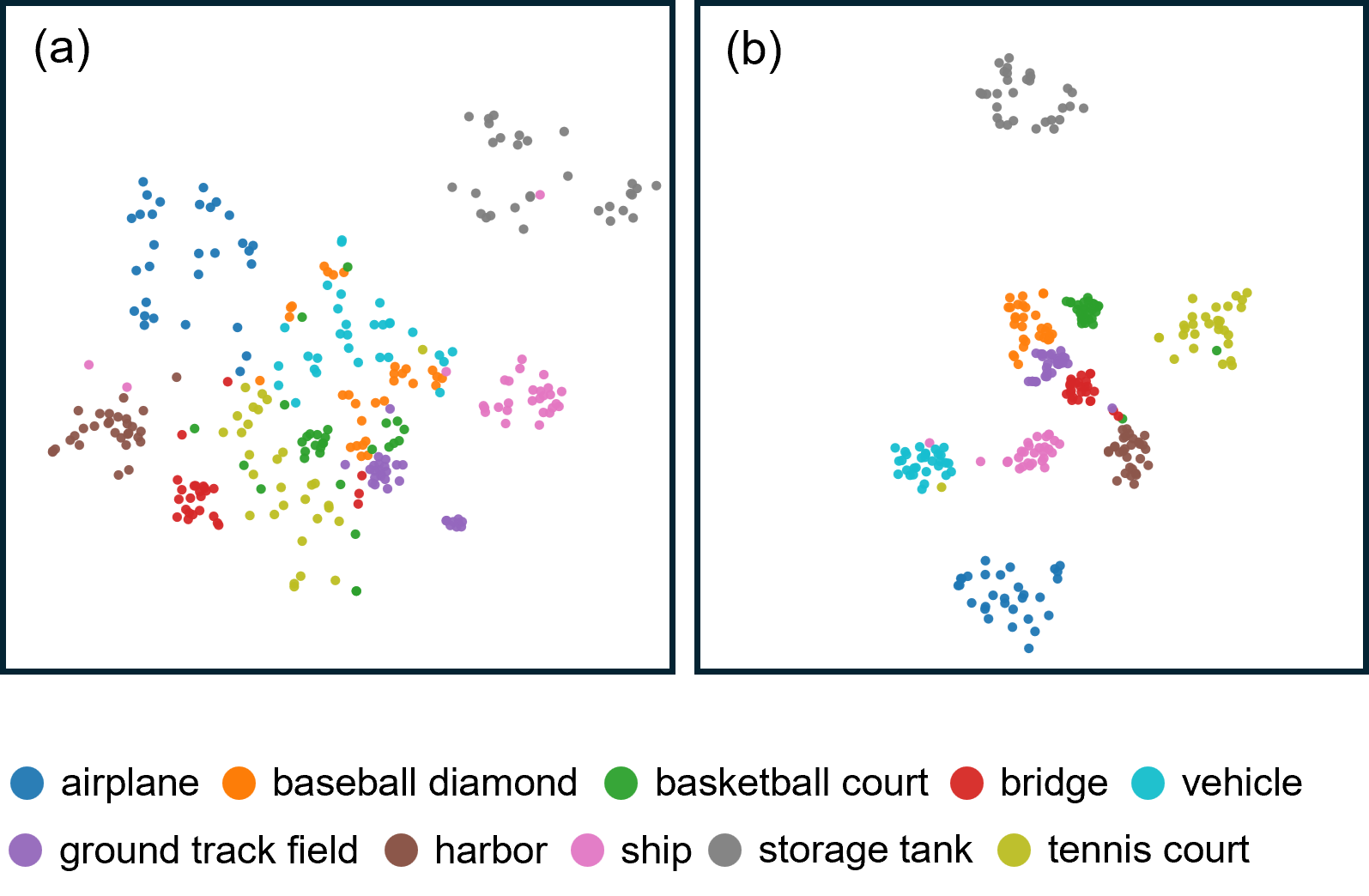}
    \caption{The t-SNE visualization of proposal embeddings learned without (a) and with (b) the HML module.}
    \label{fig6}
\end{figure}

In this subsection, we conduct ablation studies to analyze the contribution of each proposed component to the performance of HyRS-OWOD. All ablation experiments are conducted on NWPU VHR-10 under the 8+2 known--unknown class split described in Sec.~\ref{sec:split}. 

\textit{Effectiveness of DOL and HUL.}
We first evaluate the contributions of DOL and HUL to unknown object discovery. As shown in the upper part of Table~\ref{tab5}, introducing DOL alone increases U-Recall from 45.2\% to 48.1\%, indicating that decoupling class-agnostic objectness from semantic classification helps retrieve potential unknown objects. Incorporating HUL alone produces a larger improvement in U-Recall, from 45.2\% to 51.2\%, while reducing both WI and A-OSE to zero. This result demonstrates the effectiveness of the hyperbolic radius as an uncertainty cue for known--unknown discrimination. When DOL and HUL are combined, U-Recall further increases to 58.9\%, outperforming the baseline by 13.7 percentage points. Meanwhile, WI remains zero and A-OSE decreases from 8 to 1. Although K-mAP decreases moderately from 91.9\% to 89.1\%, the combined model achieves a substantially better balance between unknown-object discovery and known-class detection. These results also indicate that DOL and HUL are complementary: DOL improves class-agnostic foreground discovery, whereas HUL facilitates the subsequent discrimination between known and unknown objects.

\textit{Effectiveness of HML.}
We further investigate the effectiveness of HML during incremental learning. As reported in the lower part of Table~\ref{tab5}, adding HML improves the mAP of previously known classes from 91.5\% to 91.6\%, indicating that the model preserves the knowledge acquired in the preceding stage. More notably, the mAP of the currently introduced classes increases from 76.9\% to 91.2\%, corresponding to an improvement of 14.3 percentage points. Consequently, the overall mAP increases from 88.5\% to 91.5\%. These results suggest that HML substantially improves the learning of newly introduced classes without degrading the performance of previously known classes, thereby achieving a better balance between stability and plasticity during incremental learning.

\textit{Visualization and Analysis.} 

To further analyze the effect of HML on representation learning, we visualize the proposal embeddings learned with and without HML using t-SNE. As shown in Fig.~\ref{fig6}, without HML, embeddings from different categories exhibit substantial overlap and relatively dispersed intra-class distributions. In contrast, incorporating HML produces more compact intra-class clusters and clearer separation between different classes. This visualization demonstrates that HML improves the discriminability of proposal representations, supporting its role in reducing interference between previously learned and novel classes during incremental learning.

\section{Conclusion}

This work investigated hyperbolic geometry for OWOD in remote sensing imagery. We proposed HyRS-OWOD, which employs DOL to disentangle class-specific information from objectness. To discriminate unknown objects from known classes, we presented the HUL strategy based on the hyperbolic radius of proposal embeddings. To mitigate the forgetting effect in incremental learning, we introduced HML, which encourages the model to learn discriminative and stable embeddings by improving intra-class compactness and inter-class separability. Extensive experiments on multiple remote sensing benchmarks demonstrated that HyRS-OWOD improves unknown-object recall, reduces known--unknown confusion, and effectively retains previously acquired knowledge during incremental learning. These results validate the potential of hyperbolic representation learning for open-world object detection in remote sensing imagery. 

\bibliographystyle{IEEEtran}
\bibliography{ref}

\begin{IEEEbiography}[{\includegraphics[width=1in,height=1.25in,clip,keepaspectratio]{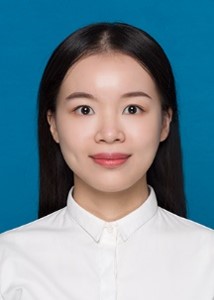}}]{Wuzhou Li}
received the B.Sc., M.Sc., and Ph.D. degrees from Wuhan University, Wuhan, China, in 2014, 2020, and 2024, respectively. She is currently a Lecturer with the School of Computer Science and Artificial Intelligence, Wuhan Textile University, Wuhan, China.

Her research interests include computer vision and deep learning, especially on object detection in remote sensing.
\end{IEEEbiography}

\begin{IEEEbiography}[{\includegraphics[width=1in,height=1.25in,clip,keepaspectratio]{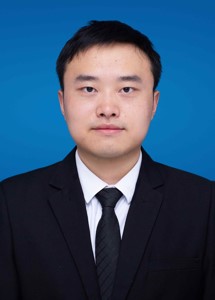}}]{Jiawei Zhou}
received the B.Sc., M.Sc., and Ph.D. degrees from Wuhan University, Wuhan, China, in 2017, 2021, and 2025, respectively. 

His research interests include deep learning, computer vision, and remote sensing image object detection.
\end{IEEEbiography}

\begin{IEEEbiography}[{\includegraphics[width=1in,height=1.25in,clip,keepaspectratio]{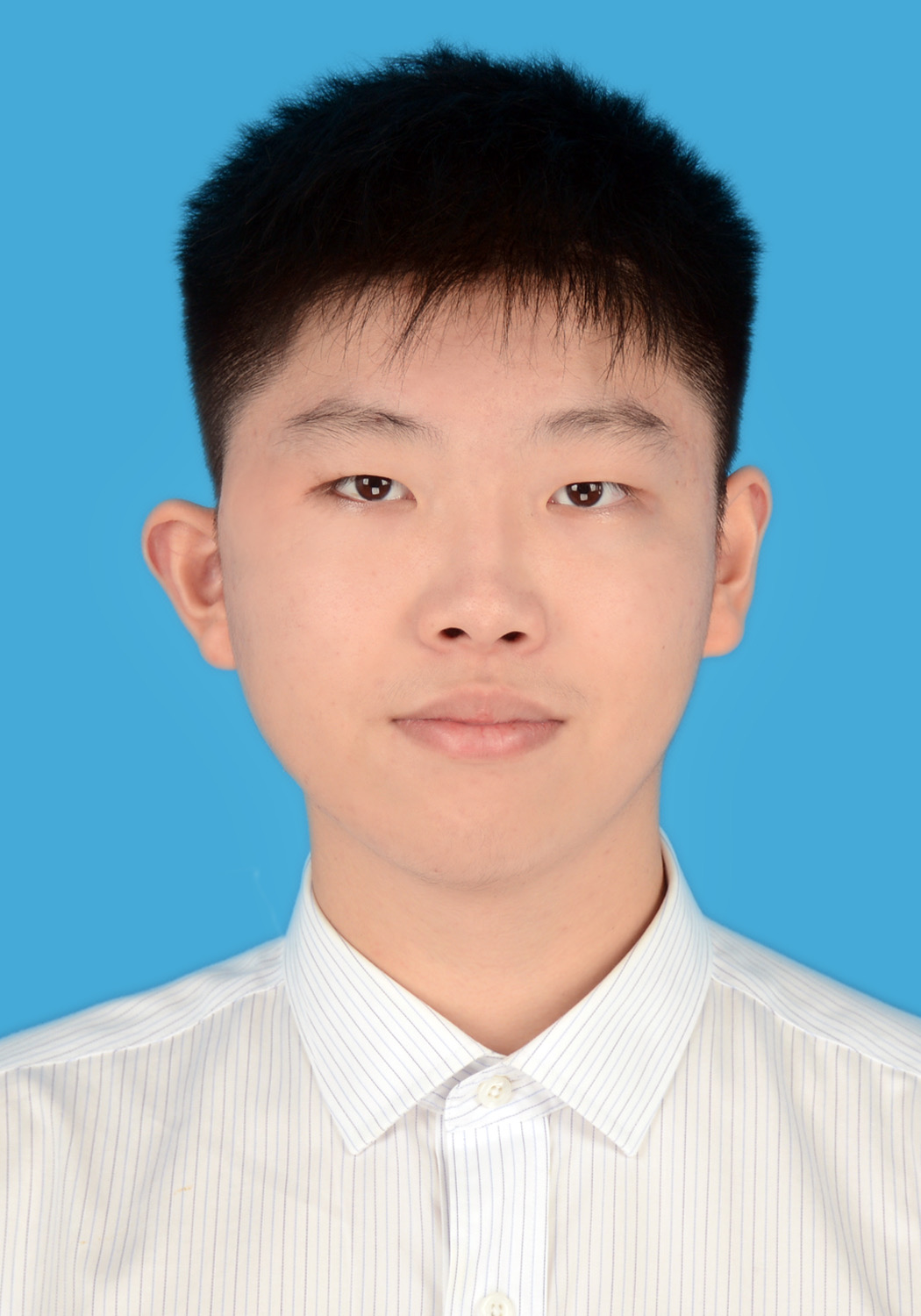}}]{Shenghang Wang}
 received the B.S. degree in Electrical Engineering Technology from Northern Kentucky University, Highland Heights, KY, USA, in 2025. He is currently pursuing the M.S. degree with the Electrical and Computer Engineering, Ohio State University, Columbus, OH, USA. 
 
 His research interests include computer vision and point cloud processing. 
\end{IEEEbiography}

\begin{IEEEbiography}[{\includegraphics[width=1in,height=1.25in,clip,keepaspectratio]{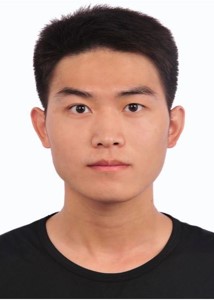}}]{Xiang Li}
received the B.Sc. degree from Wuhan University, Wuhan, China, in 2014, and the Ph.D. degree from the Institute of Remote Sensing and Digital Earth, Chinese Academy of Sciences, Beijing, China, in 2019. He is currently a Professor with with the School of Artificial Intelligence, Wuhan University, Wuhan, China. 

His research interests include deep learning, computer vision, and remote sensing image interpretation.
\end{IEEEbiography}

\end{document}